\documentclass[12pt]{article}
\usepackage[margin=1in]{geometry}

\date{}

\newcommand{\set}[1]{\left\{#1\right\}}
\newcommand{\paren}[1]{\left(#1\right)}

\usepackage{amsmath}
\usepackage{array}
\usepackage[dvipsnames]{xcolor}
\usepackage{color,hyperref}
\hypersetup{
	colorlinks,
	linkcolor={red!50!black},
	citecolor={blue!50!black},
	urlcolor={blue!80!black}
}
\usepackage{listings}
\definecolor{codegreen}{rgb}{0,0.6,0}
\definecolor{codegray}{rgb}{0.5,0.5,0.5}
\definecolor{codepurple}{rgb}{0.58,0,0.82}
\definecolor{backcolour}{rgb}{0.95,0.95,0.92}
 
\lstdefinestyle{mystyle}{
    backgroundcolor=\color{backcolour},   
    commentstyle=\color{codegreen},
    keywordstyle=\color{magenta},
    numberstyle=\tiny\color{codegray},
    stringstyle=\color{codepurple},
    basicstyle=\footnotesize,
    breakatwhitespace=false,         
    breaklines=true,                 
    captionpos=b,                    
    keepspaces=true,                 
    numbers=left,                    
    numbersep=5pt,                  
    showspaces=false,                
    showstringspaces=false,
    showtabs=false,                  
    tabsize=2
}
\usepackage{graphicx}
\graphicspath{ {} }

\usepackage{soul}  
\usepackage[normalem]{ulem}  
\newcommand{\Fworkname}{\texttt{JANOS}}

\definecolor{tengscolor}{rgb}{0.0, 0.0, 1.0}  

\usepackage{authblk}
\usepackage{natbib}
 \bibpunct[, ]{(}{)}{,}{a}{}{,}%
\title{$\Fworkname$ : An Integrated Predictive and Prescriptive Modeling Framework}
\author[1]{\small David Bergman}
\author[1]{\small Teng Huang}
\author[2]{\small Philip Brooks}
\author[3]{\small Andrea Lodi}
\author[4]{\small Arvind U. Raghunathan}
\affil[1]{\footnotesize Department of Operations and Information Management, School of Business, University of Connecticut, United States, \{david.bergman, teng.huang\}@uconn.edu}
\affil[2]{\footnotesize Optimized Operations, LLC., phil.brooks24@me.com}
\affil[3]{\footnotesize Department of Mathematical and Industrial Engineering, Polytechnique Montreal, Canada, andrea.lodi@polymtl.ca}
\affil[4]{\footnotesize Mitsubishi Electric Research Laboratories, raghunathan@merl.com}

\begin{document}






\maketitle
\begin{abstract}
Business research practice is witnessing a surge in the integration of predictive modeling and prescriptive analysis.  
We describe a modeling framework $\Fworkname$
that seamlessly integrates the two streams of analytics, for the first time allowing researchers and practitioners to embed machine learning models in an optimization framework.
$\Fworkname$ allows for specifying a prescriptive model using standard optimization modeling elements such as constraints and variables.  The key novelty lies in providing modeling constructs that allow for the specification of commonly used predictive models and their features as constraints and variables in the optimization model. 
The framework considers two sets of decision variables; \emph{regular} and \emph{predicted}.  The relationship between the regular and the predicted variables are
specified by the user as pre-trained predictive models.  $\Fworkname$ currently supports linear regression, logistic regression, and neural network with rectified linear activation functions, but we plan to expand on this set in the future. 
In this paper, we demonstrate the flexibility of the framework through an example on scholarship allocation in a student enrollment problem and provide a numeric performance evaluation. 
\end{abstract}



%


\section{Introduction}
\label{sec:intro}

There has been significant proliferation of research in and application of machine learning and discrete optimization.  
These two analytical domains have frequently been used in a single business decision-making problem but for different purposes. Machine learning techniques have typically been used to predict what is likely to happen in the future, while optimization methods have been used to strategically search through feasible solutions.

However, the integration of the two analytics streams is often disjointed.  
As an example, suppose a neural network is built to predict customer churn of a telecommunication company, with features consisting of demographic information together with service 
price
to the customer (the lower the price, the lower the probability of customer churn).  
How then should the company set service price in order to maximize revenue?  The features of the predictive model are variables in the decision problem. 
In other words, the output of neural networks whose features (e.g., service price) are decision variables are part of the objective function.  This makes the optimization problem of maximizing revenue particularly challenging.  There are few, if any, techniques, much less tools, available for solving such an optimization problem. This paper seeks to fill that void by introducing $\Fworkname$, a modeling framework for joint predictive-and-prescriptive analytics. 



$\Fworkname$ allows a user to specify portions of an objective function as commonly utilized predictive models, whose features are fixed or are variables in the optimization model.  $\Fworkname$ currently supports predictive models of the following three forms: linear regression, logistic regression, and neural networks (NNs) with rectified linear (ReLU) activation functions.  These models are commonly used in practice, and the framework is easily extensible to other predictive models.

To embed these predictive models in an optimization model, we utilized linearization techniques. For linear regression this is straightforward.  
For logistic regression, we employ a piece-wise linear approximation. Details are provided in Section~\ref{subsec:Log}.
For NNs, we make use of recent work that formulates a NN as a \emph{mixed integer programming} (MIP) problem \citep{serra2017bounding,bienstock2018principled,fischetti2017deep}; we do not use the MIP to learn the NN, but rather utilize the network reformulation to produce outputs of the NN based on the input features. Details are provided in Section~\ref{subsec:NN}.

A key advantage of $\Fworkname$ is that it automates the transcription of common predictive models into constraints that can be handled by mixed integer programming solvers. Thus, researchers and practitioners are relieved of this onerous task of reformulating the predictive models into tractable constraints.  A different model for predictive variables can quickly be substituted out without much effort from the modeler, enabling the user to quickly compare the optimal decisions when different predictive models are used. Additionally, one need not worry about modeling all parameters of a predictive model with an optimization model; for example, a neural network with 3 layers of 10 nodes each would have hundreds of parameters, and $\Fworkname$ automates that transformation. 
In future releases, additional predictive models will be added as well as more advanced reformulations and algorithmic implementations.

The framework, which we call $\Fworkname$\footnote{A play on Janus, who according to ancient Roman mythology is the god of beginnings, gates, transitions, time, duality, doorways and is usually depicted with two faces one looking to the past (predictive) and one to the future (prescriptive). (Source: Wikipedia)}, is built in \texttt{Python} and calls \texttt{Gurobi} \citep{gurobi} to solve MIPs.   A user specifies an optimization model through standard modeling constructs that share similarities to those in other common optimization modeling languages, for example \texttt{Gurobi}'s \texttt{Python} interface, \texttt{Pyomo}~\citep{hart2011pyomo} or \texttt{Julia} \citep{bezanson2017julia}. 

We partition the variables in the model into two sets\textemdash the \emph{regular variables} and the \emph{predicted variables}.  The regular variables are used to model operational constraints and objective function terms as typical variables in MIP.  The predicted variables are specified through predictive models wherein some of the features depend on regular variables.  
The predictive models are pre-trained by the user, 
who can load any of the three permissible predictive model forms, together with a \emph{mapping} between the regular variables and the features.  
We eventually plan to integrate automated machine learning~\citep{feurer2015efficient} and have $\Fworkname$ determine the best predictive model to associate with a given data frame.
To exhibit how the framework can be used, we present as an example the allocation of scholarship offers to admitted students to optimize the enrollment.

The rest of the paper is organized as follows. We first review the literature related to the joint reasoning in predictive and prescriptive analysis in Section~\ref{sec:litReview}. 
The general problem addressed by $\Fworkname$ is provided in Section~\ref{sec:probDesc}. 
The algorithmic details of how we optimize over linear regression models, logistic regression models and NNs are given in Section~\ref{sec:algoDetails}.
The student enrollment example and a collection of experiments designed to test the efficiency of  $\Fworkname$ are described in Section~\ref{sec:examApp}. 
We then describe how $\Fworkname$ can be downloaded and installed in Section~\ref{sec:solveraccess}.  
We conclude in Section~\ref{sec:conclusion}.

\section{Literature Review}
\label{sec:litReview}

Existing studies on the combination of predictive and prescriptive analytics take predictions as fixed and then make choices based on fixed predictions, for example, predictions are parameters in an optimization model \citep{ferreira2015analytics}.
\cite{ferreira2015analytics} first predict sales based on a chosen number of price values and then uses those fixed estimates to determine the optimal price. 
In their application, the predicted sales are parameters in the optimization model. This modeling approach adapts, but still lacks in full integration between the two analytics disciplines.  Sales will generally depend on price, and analytical methods by which one can optimize decisions that ultimately affect objective functions are currently lacking in the literature. 
Approaches using fixed-point estimates of parameters are feasible when full enumeration or partial enumeration of the collection of feasible solutions is practical.
However, in instances where enumeration is not possible, facets of the optimization algorithm need to be integrated into the predictive model.  For example,
\cite{huang2019predictive} model a real-world problem in such a way, but only propose an exact optimization approach when simple linear regression models are used for prediction.

There are additional streams of research that combine predictive modeling and optimization.
First of all, machine learning algorithms are powered by optimization techniques. For example, when fitting a simple linear regression model, the \emph{ordinary least square} method determines the unknown parameters by minimizing the sum of the squares of the difference between the observed dependent variable values and the fitted values.
In machine learning, various optimization techniques are applied so that the learning process is efficient and achieves desired accuracy \citep{boyd2011distributed}.
For example,
\cite{koh2007interior} propose an interior-point method for solving large-scale $\ell_1$-regularized logistic regression problems;
\texttt{ALAMO} \citep{cozad2014learning} uses mixed integer optimization to learn algebraic models from data; 
and, linear programming (LP) and integer programming (IP) based methods have been proposed to assist training NNs \citep{serra2017bounding,bienstock2018principled,fischetti2017deep}.
Additionally, there have been recent efforts on using machine learning to improve optimization algorithms. For example, \cite{cappart2019improving} utilize deep reinforcement learning to improve optimization bounds, and \cite{khalil2017learning} proposes a generic method for learning combinatorial optimization over graphs, with a plethora of papers arising in this area \citep{nazari2018reinforcement,lemos2019graph,bengio2018machine}.  These papers either leverage machine learning to improve optimization, or leverage optimization to improve machine learning\textemdash in our setting, we integrate the two into a unified decision-making framework. 

The impact of the $\Fworkname$ solver on research is twofold.  First, framing a problem as optimizing over standard predictive models where the decision variables are features of the predictive models requires much attention. Suppose the predictive model is a support vector regression with a radial kernel.  How does one solve the optimization problem?  Each predictive model added to $\Fworkname$ will require research effort to investigate linearizations or advanced optimization methodology that lead to efficient solution times. 

More broadly, we envision that $\Fworkname$ will be used by researchers in fields other than optimization to solve the problems they are interested in to derive the insights they desire.  Currently, a researcher familiar with optimization who has a machine learning model that they would like to optimize over is unable to solve the problem of interest.  Contrast that with a researcher in logistics that requires the solution of a routing problem to identify how many trucks are needed by the company.   This researcher can use a standard commercial integer programming solver, solve the routing problem to identify the number of trucks that are needed by the company, and then make decisions based on that output.  The logistics researcher need not know how the optimization model is solved, just that the problem of interest can be solved.  We envision the same thing for the former researcher.  $\Fworkname$ allows the researcher to solve an optimization problem with machine learning models so that optimal decisions can be identified for real-world decision making problems that previously could not. 

\section{Problem Description}
\label{sec:probDesc}

$\Fworkname$ seeks to solve problems formulated as (\ref{prob:ori}):
\begin{align}
\label{prob:ori}
\tag{\textbf{PROBLEM-ORI}}
& \max\limits_{x} && \sum_{j=1}^{n_1} c_j x_j + \sum_{k=1}^{n_2} d_k y_k \nonumber\\
& \text{s.t.} && \sum_{j=1}^{n_1} a_j^i x_j \leq b_i, && \forall i \in \set{1, \ldots, m}\\
&&& y_k = g_k(\alpha^k_{1}, \ldots, \alpha^k_{p_k}; \theta_k), && \forall k \in \set{1, \ldots, n_2}\\
&&& \alpha^k_{l} \text{ is given}, && \forall l \in \set{1,\ldots,q_k}, \forall k \in \set{1, \ldots, n_2}\\
&&& \alpha^k_{l} = e^k_l \cdot x, && \forall l \in \set{(q_k+1),\ldots,p_k}, \forall k \in \set{1, \ldots, n_2}\\ 
&&& x_j \in X_j, && \forall j \in \set{1, \ldots, n_1}.
\end{align}
The variables
$x = (x_1, \ldots, x_{n_1})$ are \emph{regular variables} and the variables
$y = (y_1,\ldots, y_{n_2})$ are \emph{predicted variables}.  Each variable $x_j$ belongs to a finite or continuous set $X_j$ and are constrained via linear inequalities.  Each predicted variable $y_k$ is associated with a predictive model $g_k$, with features  $\alpha^k=(\alpha^k_1,\ldots,\alpha^k_{p_k})$. Each $g_k$ is assumed to be a pre-trained predictive model (a linear regression, logistic regression, or neural network with a rectified linear activation function)  so that the parameters $\theta_k$ are fit prior to optimization by the user.  Model $g_k$ has $p_k$  \emph{features}, 
 $(\alpha^k_1,\ldots,\alpha^k_{p_k})$. 
The first $q_k, 1 \leq q_k \leq p_k$, features of predictive model $g_k$ are fixed and given, while each of the remaining $ (p_k - q_k)$ features are regular variables, linked through the function $e^k_l$, a $n_1$-length binary unit-vector with a 1 in the coordinate of the associated regular variable.  Note that a single pre-trained model can be used as multiple $g_k$'s. 


\section{Algorithmic Details}
\label{sec:algoDetails}

In this section, we summarize how $\Fworkname$ handles linear regression, logistic regression, and NN models.
If  $y_k$ is determined by a linear regression model, the function $g$ is linear and the construction is straightforward. We construct~(\ref{prob:ori}) and feed the model to \texttt{Gurobi}.
If $y_k$ is a predicted value of a NN, $y_k$ is obtained from a network flow model. The details are in Section~\ref{subsec:NN}.
If $y_k$ is a predicted value of a logistic regression model, we partition the range of the \emph{log-odds}, i.e., $X\beta$ as in $y_k = \frac{1}{1+e^{-X\beta}}$ into a collection of intervals, and use the mean of the $\mathbf{sigmoid}$ value of $X\beta$ within the corresponding interval to approximate $y_k$ in the objective function. The details are in Section~\ref{subsec:Log}.

\subsection{Optimization over Neural Networks}
\label{subsec:NN}

For every predicted variable $y_k$ that is determined by a NN prediction, we associate a distinct network flow model. A NN can be viewed as a multi-source single-terminal\footnote{The NN does not always have a single terminal. In our case, we only have one output, and so the output layer in our trained NN only has one node.} arc-weighted acyclic layered digraph $\mathsf{N} = (\mathsf{V},\mathsf{A})$.  
The node set $\mathsf{V}$ is partitioned into a collection of layers $\mathsf{V_1} \cup \cdots \cup \mathsf{V}_l$. There is a one-to-one mapping between input features $\alpha$ of $g$ and nodes $\mathsf{v} \in \mathsf{V}_1$.  
For any node $\mathsf{v} \in \mathsf{V}_1$, we denote the corresponding feature by $\alpha(\mathsf{v})$.
Set $\mathsf{V}_l$ consists of a single terminal node $\mathsf{t}$.  For $j \in \set{ 2, \ldots, l}$, every node $\mathsf{u} \in \mathsf{V_j}$ has a \emph{bias} $\mathsf{B(u)}$ learned during training. 

Each arc $\mathsf{a} = (\mathsf{u},\mathsf{v}) \in \mathsf{A}$ is directed from a node $\mathsf{u}$ in layer $\mathsf{V}_j$ to a node $\mathsf{v}$ in layer $\mathsf{V}_{j+1}$ for some $j \in \set{1, \ldots, l-1}$.  Every arc has a \emph{weight} $\mathsf{w}(\mathsf{a})$ learned during training.  

Given values $\alpha(\mathsf{v})$ for all nodes $\mathsf{v} \in \mathsf{V}_1$, the prediction from a NN with a ReLU activation function is calculated recursively by assigning a value $\mathsf{F}_\mathsf{v}$ to all nodes in the NN via the following iterative procedure and returning $\mathsf{F}_\mathsf{t}$:
\begin{itemize}
    \item $\forall \mathsf{v} \in \mathsf{V}_1, \mathsf{F}_\mathsf{v} = \mathsf{G}_\mathsf{v} =  \alpha(\mathsf{v})$;
    \item  For $j = 2, \ldots, l-1, \forall \mathsf{v} \in \mathsf{V}_{j}$, 
    $\mathsf{G}_\mathsf{v} = \sum_{\mathsf{u} \in \mathsf{V}_{j-1}} \mathsf{w}\paren{\paren{\mathsf{u},\mathsf{v}}} \cdot \mathsf{F}_\mathsf{u} + \mathsf{B}(\mathsf{v})$, 
    $\mathsf{F}_\mathsf{v} = 
    \max \set{0, \mathsf{G}_\mathsf{v}}$; and
    \item $\mathsf{F}_\mathsf{t} = \mathsf{G}_\mathsf{t} = \sum_{\mathsf{u} \in \mathsf{V}_{l-1}} \mathsf{w} \paren{\paren{\mathsf{u},\mathsf{t}}} \cdot \mathsf{F}_\mathsf{u} + \mathsf{B}(\mathsf{u})$.
\end{itemize}
Here  $\mathsf{G}_\mathsf{v}$ is the input of the ReLU function, and
    $\mathsf{F}_\mathsf{v}$ is the output of the ReLU function.

We further define $z_{\mathsf{u}}$ $\forall \mathsf{u} \in \mathsf{V}$ as a binary variable indicating if $\mathsf{G}_{\mathsf{u}} >0$.
With this interpretation, one can formulate the following model (\ref{mod:nn}) to calculate $y_k$ based on the inputs $\alpha(\mathsf{v})$, $\forall \mathsf{v} \in \mathsf{V}_1$, that are the features, which can be fixed constants or functions of the decision variables.  
\begin{align}
\tag{\textbf{MOD-NN}}
&&& y_k = \mathsf{F}_\mathsf{t} && \label{mod:nn}  \\
&&&
\mathsf{F}_{\mathsf{v}} = \mathsf{G}_{\mathsf{v}}, 
    && \forall \mathsf{v} \in \mathsf{V}_1 \cup \mathsf{V}_l \label{constr:preequalpost}\\
&&& \mathsf{G}_{\mathsf{v}} = \alpha(\mathsf{v}),
    && \forall \mathsf{v} \in \mathsf{V}_1
    \label{constr:inputlayer}\\
&&& \mathsf{G}_{\mathsf{v}} = \sum_{\mathsf{u} \in \mathsf{V}_{j-1}} \mathsf{w}(\paren{\mathsf{u}, \mathsf{v}}) \cdot \mathsf{F}_{\mathsf{u}} + \mathsf{B}(\mathsf{u}), 
    && \forall j \in \set{2, \ldots, l}, \forall \mathsf{v} \in \mathsf{V}_j \label{constr:nnlayers} \\
&&& -M \cdot \paren{1-z_\mathsf{v}} \leq \mathsf{G}_\mathsf{v} \leq M \cdot z_\mathsf{v},  
    && \forall \mathsf{v} \in \mathsf{V}_2 \cup \cdots \cup \mathsf{V}_{l-1} \label{constr:beginrelu} \\
&&& \mathsf{G}_\mathsf{v} - M \paren{1-z_\mathsf{v}} \leq \mathsf{F}_\mathsf{v} \leq  \mathsf{G}_\mathsf{v} + M \cdot \paren{1 - z_\mathsf{v}}, 
    && \forall \mathsf{v} \in \mathsf{V}_2 \cup \cdots \cup \mathsf{V}_{l-1} \\
&&& 0 \leq \mathsf{F}_{\mathsf{v}} \leq M \cdot z_\mathsf{v}, 
    && \forall \mathsf{v} \in \mathsf{V}_2 \cup \cdots \cup \mathsf{V}_{l-1} \label{constr:endrelu} \\
&&& y_{\mathsf{v}} \in \set{0, 1}, 
    && \forall \mathsf{v} \in \mathsf{V}\\
&&& \mathsf{G}_\mathsf{v} \text{ unconstrained}, 
    && \forall \mathsf{v} \in \mathsf{V}\\
&&& \mathsf{F}_\mathsf{v} \text{ unconstrained}, 
    && \forall \mathsf{v} \in \mathsf{V}.
\end{align}
Constraints (\ref{constr:preequalpost}) guarantee that the ReLU activation function is not enforced in the input and output layers.
Constraints (\ref{constr:inputlayer}) enforce that the values of the nodes in the input layer is the values of each input variable of the predictive model.  These will either be fixed constants or the value determined by the optimization model for a single problem variable. 
Constraints (\ref{constr:nnlayers}) compute the input of the ReLU function of each node that is not on the first layer.
Constraints (\ref{constr:beginrelu}) to (\ref{constr:endrelu}) enforce the ReLU activation function on each hidden layer.

In~(\ref{prob:ori}), $\alpha(v)$ can be a regular variable or a constant, and there will be a separate network flow formulation for each of the predicted variables that are outcomes of neural networks. 
We test the impact of the size of the neural network and the number of such predicted variables on the performance of $\Fworkname$ in Section~\ref{sec:examApp}.

\subsection{Optimization over Logistic Regression Models}
\label{subsec:Log}

\begin{figure}[!htbp] 
	\centering
	\includegraphics[scale=0.13]{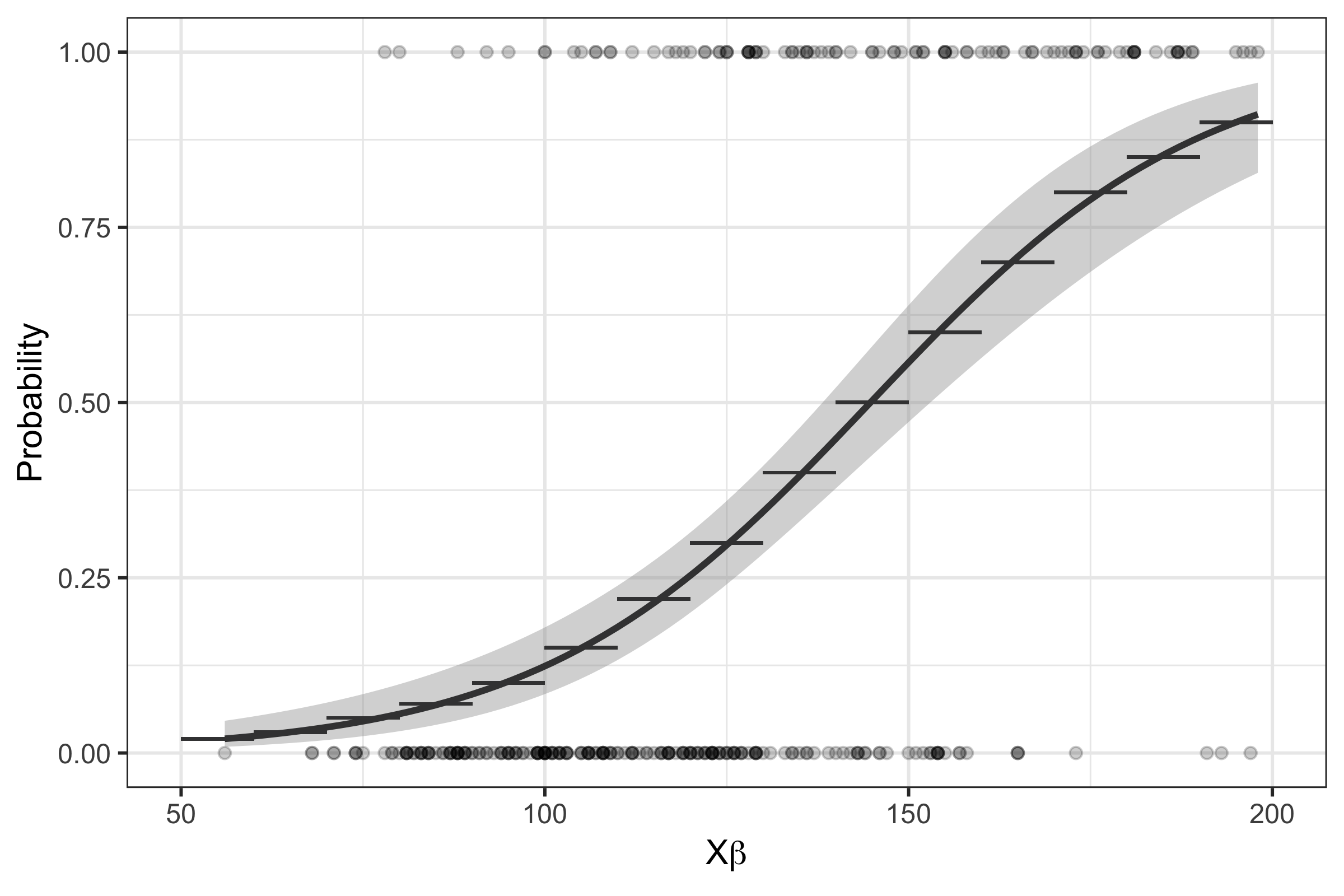} 
	\caption{A logistic function curve.} 
	\label{fig:logit} 	
\end{figure}

$\Fworkname$ provides a parameterized discretization for handling logistic regression prediction.
Specifically, it represents the nonlinear function of a logistic regression model using a piece-wise linear approximation, partitioning the range of the \emph{log-odds}, i.e., $X\beta$ as in $y_k = \frac{1}{1+e^{-X\beta}}$ into a collection of mutually exclusive and collectively exhaustive intervals. 
The number of intervals is a parameter that can be specified by users.
This idea is illustrated on Figure~\ref{fig:logit}.

We partition the range of the \emph{log-odds} into $\Delta$ intervals, $[L_\delta, U_\delta]$, for $\delta \in \set{1, \ldots, \Delta}$, where $L_{\delta + 1} = U_{\delta}$ and we assume that the length of each interval is uniform. 
The range of the \emph{log-odds} is computed based on the bounds of the features.
We use the mean of the $\mathbf{sigmoid}$ function value within an interval to serve as a piece-wise linear approximation of the actual predicted value of the logistic regression model.
Specifically, for interval $\delta$, let 
    \[
    \mathcal{V}_\delta = \frac{\log(1+e^{U_{\delta}}) - \log(1+e^{L_\delta})}{U_{\delta} - L_{\delta}}.
    \]
The value $\mathcal{V}_{\delta}$ is the average value of the $\mathbf{sigmoid}$ function over all values between $L_\delta$ and $U_\delta$, where
\[
\mathbf{sigmoid}(\mathbf{a}) = \frac{e^\textbf{a}}{1+e^\textbf{a}}.
\]

We define $z_\delta$ $\forall  \delta \in \set{1, \ldots, \Delta}$ as a binary variable indicating if we select a value for $y_k$ in interval $\delta$.
Let $F$ be the vector of features.
Let $\beta$ be the vector of estimated coefficients in the logistic regression model.
With this interpretation, one can formulate (\ref{mod:log}) to maximize over a logistic regression model \emph{approximately}, using the following transformation:
\begin{align}
\label{mod:log}
\tag{\textbf{MOD-LOGIT}}
& \text{max} && y_k \nonumber\\
& \text{s.t.} && \sum_{\delta=1}^{\Delta} z_\delta = 1, \label{constr:onez}\\
&&& (L_\delta - L_1) \cdot z_\delta + L_1 \leq F \beta, && \forall \delta \in \set{1, \ldots, \Delta} \label{constr:originvaluebegin}\\
&&& (U_{\delta} - U_\Delta) \cdot z_\delta + U_\Delta \geq F \beta, && \forall \delta \in \set{1, \ldots, \Delta} \label{constr:originvalieend}\\
&&& \mathbf{sigmoid}(L_1) + (\mathcal{V}_\delta - \mathbf{sigmoid}(L_1)) \cdot z_\delta \leq y_k,
    && \forall \delta \in \set{1, \ldots, \Delta} \label{constr:beginpredict} \\
&&& \mathbf{sigmoid}(U_\delta) + (\mathcal{V}_\delta - \mathbf{sigmoid}(U_\delta)) \cdot z_\delta \geq y_k,
    && \forall \delta \in \set{1, \ldots, \Delta} \label{constr:endpredict}\\
&&& z_\delta \in \set{0, 1}. && \forall \delta \in \set{1, \ldots, \Delta}
\end{align}
The value assumed by $y_k$ will be approximately equal to $\mathbf{sigmoid}(F\beta)$, used as the value predicted by a logisitic regression model. 

Constraint (\ref{constr:onez}) ensures that only one interval is selected.
Constraints (\ref{constr:originvaluebegin}) to (\ref{constr:originvalieend}) select the interval that contains the linear combination $F\beta$.
Constraints (\ref{constr:beginpredict}) to (\ref{constr:endpredict}) make sure that $y_k$ equals the mean outcome value of the selected interval. Recall that $F$ will be determined partially through fixed features and partially through decision variables. 

\section{Example Applications}
\label{sec:examApp}

In this section, we explore an example of allocating offered scholarship to admitted college students to exhibit the capability and flexibility of $\Fworkname$. All predictive models were built in \texttt{Python3.7} using \texttt{scikit-learn 0.21.3} \citep{scikit-learn} and all optimization models were solved with \texttt{Gurobi Optimizer v8.0} \citep{gurobi}.  All experiments were run in \texttt{MacOS Mojave 10.14.5} on a 2.8 GHz Intel Core \texttt{i7-4980HQ} processor with 16 GB RAM.  

\subsection{Problem Description}
The Admission Office of a university wants to offer scholarship to its admitted students in order to bolster the class profile, such as academic metric, and often to simply maximize the expected class size \citep{maltz2007decision}.

The Admission Office has collected from previous enrollment years the applicants' SAT, GPA, scholarship offer, and matriculation result, i.e., whether the student accepted the offer or not.
This year, suppose the school is issuing $N$ offers of admission. Moreover, suppose the budget available for offered scholarship is \$$0.2 \cdot N \cdot 10^4$, denoted by \textbf{BUDGET} henceforth.
The amount of scholarship that can be assigned to any particular applicant is between \$0 and \$25,000. The Admission Office wants to maximize the incoming class size.

To solve this optimization problem using $\Fworkname$, one can pre-train a model to predict the probability of a candidate accepting an offer given this student's SAT, GPA and scholarship offered.  The decision to make is the third feature \emph{for each student}: the amount of scholarship to offer to each student.

We model this problem as follows (\ref{prob:studEnroll}):
\begin{align}
\label{prob:studEnroll}
\tag{\textbf{STUDENT-ENROLL}}
& \text{max} && \sum_{i = 1}^{N} 1 \cdot y_i \nonumber\\
& \text{s.t.} && \sum_{i=1}^{N} x_i \leq \textbf{BUDGET}\\
&&& y_i = g(\mathbf{s}_i,\mathbf{g}_i,x_i;\theta), && \forall i \in \set{1, \ldots, N} \\
&&& 0 \leq x_i \leq 25,000, && \forall i \in \set{1, \ldots, N},
\end{align}
where,
\begin{itemize}
    \item $x_i$ is the decision variable, i.e., the amount of  scholarship assigned to each student accepted;
    \item $\mathbf{s}_i$ is the SAT score of applicant $i$ (standardized using $z$-score);
    \item $\mathbf{g}_i$ is the GPA score of applicant $i$ (standardized using $z$-score);
    \item $y_i$ is a predicted variable per admitted student, the outcome of a predicted model representing the probability of a candidate accepting the offer; and
    \item $g$ is a predicted model pre-trained to predict any candidates' probabilities of accepting an offer. The parameters $\mathbf{s}_i$, $\mathbf{g}_i$ and $x_i$ are the predictive model's inputs. The vector $\theta$ represents the parameters of the predictive model, which we assume to be the same for each applicant. The function $g$ can be any of the permissible predictive models with $\theta$ determined prior to optimization. 
\end{itemize}


\subsection{Experimental Results}

We utilize randomly generated realistic student records to train predictive models and test the efficiency of the solver when solving different-sized problems with variations in parameters as well.  
We build the three permissible models (linear regression, logistic regression, and neural networks) with various parameter settings, i.e., the number of intervals for logistic regression models and the hidden layer sizes for neural networks.  Each of the models is trained on 20,000 randomly generated student records.

We then generate sets of random student records of different sizes to test the scalability of $\Fworkname$,  50, 100, 500 and 1,000 admitted students.
These experiment instances are produced using the same data-generating scheme as was used for building the training set.
We document how long it takes $\Fworkname$ to solve problems of different sizes with different predictive models.

We first provide an analysis of the total runtime using various parameters for each of the models.  For the linear regression there are not configurable parameters, and so we have only one setting, \textbf{LinReg}.  For the logistic regression model, the main parameter of interest is the number of intervals in the discretization. For a fixed number of intervals $\Delta$, let \textbf{LogReg}$(\Delta)$ refer to solving the model with logistic regression prediction with $\Delta$ intervals.  For neural networks, there are several parameters one might tune, most apparent being the configuration of the neurons.  We fix 10 nodes per hidden layer and vary the number of hidden layers to 1, 2, 3.  Let \textbf{NN}$(h)$ refer to solving the model with neural networks with $h$ hidden layers.

\begin{figure}[!htbp] 
\centering
\includegraphics[scale=0.12]{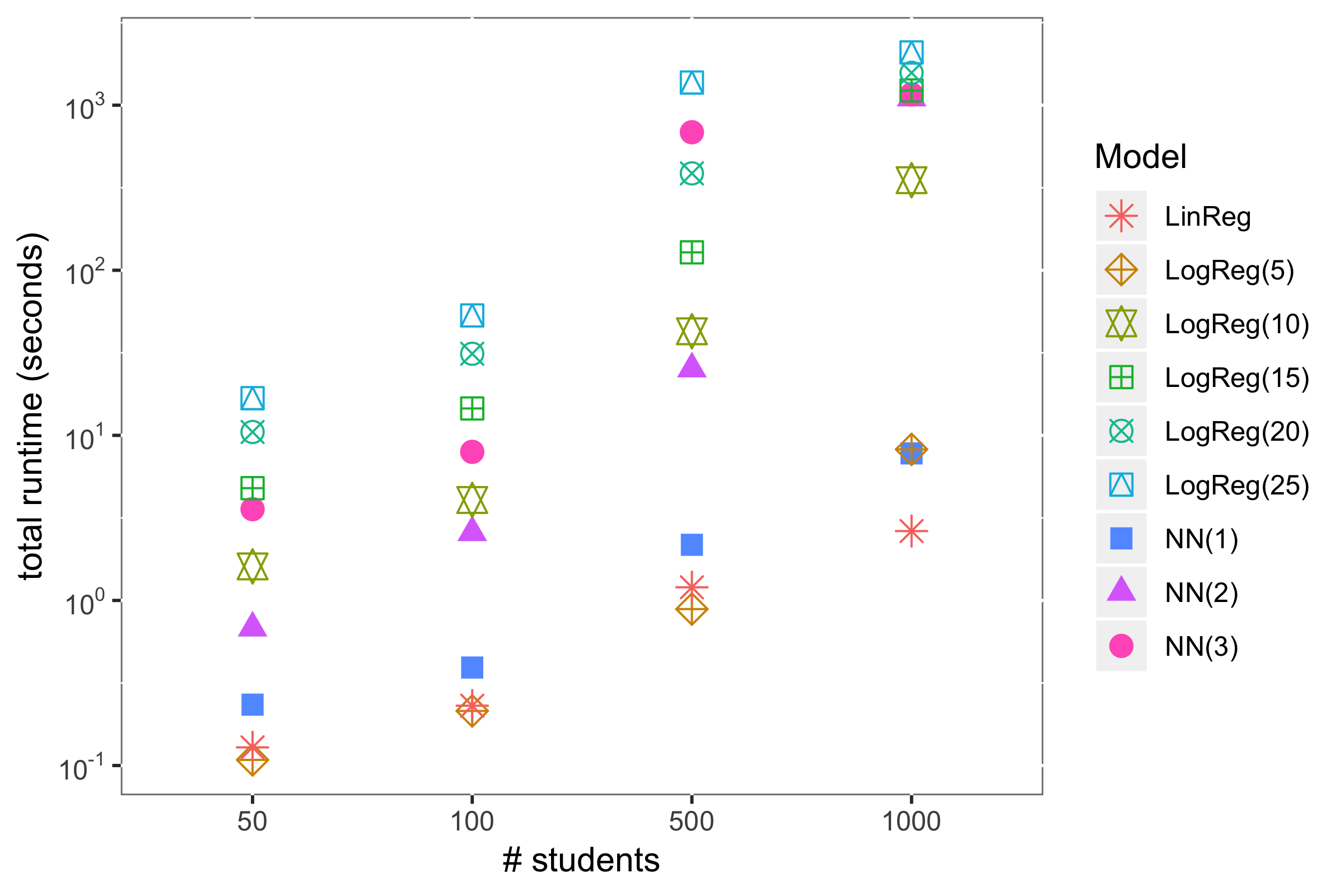}
\caption{The average runtimes of linear regression models, logistic regression models, and neural networks with different scales.} 
\label{fig:janosPerf} 	
\end{figure}

For each of the predictive model specifications, we run 5 instances and take the mean runtime.  Figure~\ref{fig:janosPerf} depicts the runtimes.  On the $x$-axis we indicate the number of admitted students in the admitted pool. On the $y$-axis we report the runtime in seconds.  \textbf{LinReg} yields the most efficient model, taking up to a second to solve.  \textbf{LogReg}($\Delta$) takes an increasing amount of time as $\Delta$ grows, as does \textbf{NN}($h$) for increasing $h$, but  the runtimes are not prohibitively large.  Note that largest instances have 1,000 logistic regression approximations or 1,000 neural network flow models.   

\begin{figure}[!htbp] 
\centering
\includegraphics[scale=0.11]{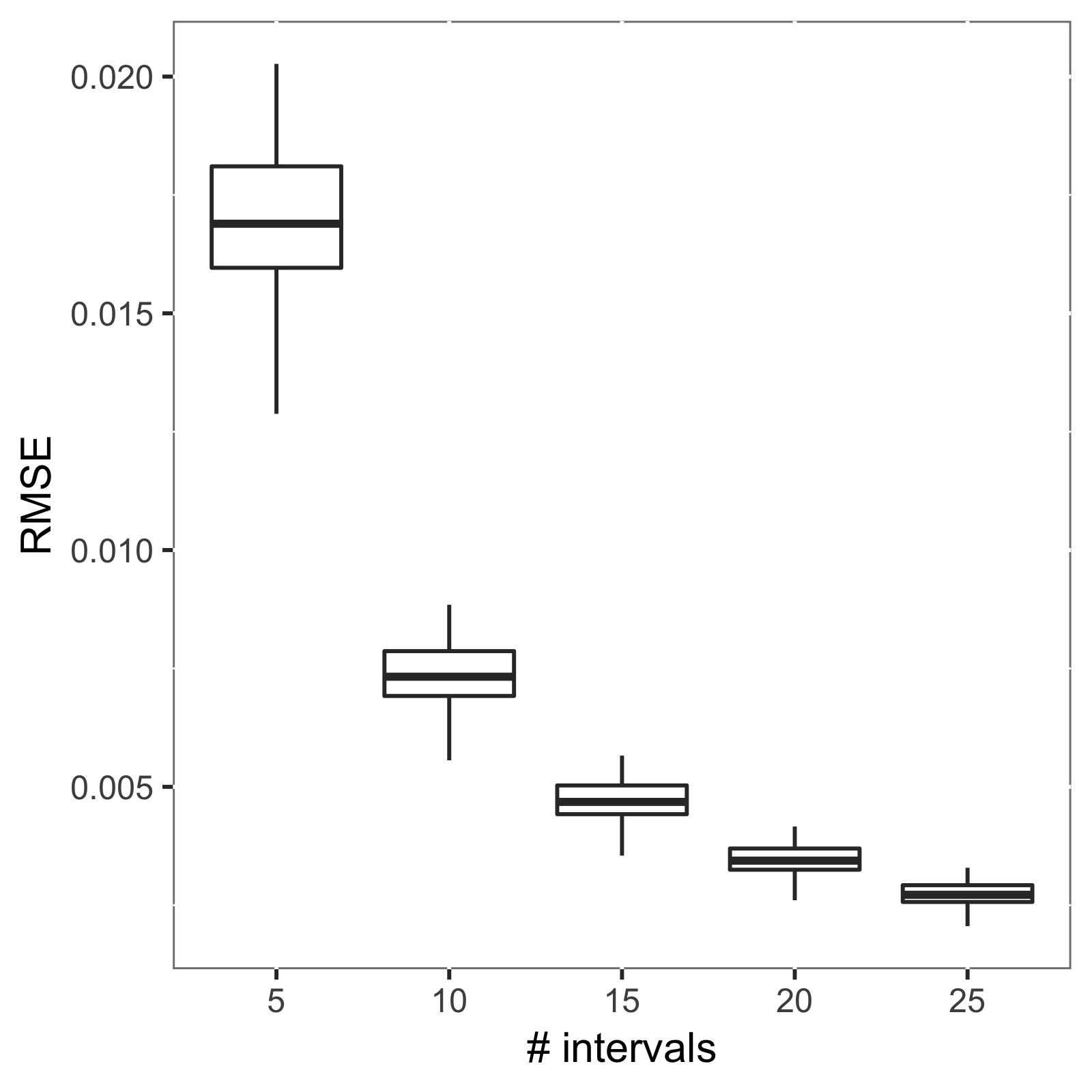}
\caption{The quality of the linear approximation of the logistic regression function at optimal solutions.} 
\label{fig:logRegPerform} 	
\end{figure}

We also evaluate how well the approximation of the logistic regression performs at obtain optimal solutions.  We apply the logistic regression approximation for 10 instances with 50 students.  The results are reported on Figure~\ref{fig:logRegPerform}. On the $x$-axis we indicate the number of intervals in the approximation, and on the $y$-axis we report the \emph{root mean squared error} (RMSE) of the probability estimates given by the approximation and the actual learned logistic regression evaluation at the optimal solutions obtained by the approximation.  As the number of intervals increases, the approximation becomes stronger, but as discussed earlier, increases runtime. Note that even with only 5 intervals, the average error in the estimate of the probability of enrollment in the approximation built is less than 0.02, or 2\%.

In order to evaluate the expected improvement in solutions obtained from $\Fworkname$ over what might be done in practice, we evaluate the solution obtained by $\Fworkname$ and compare with the following heuristic that can be employed for any predictive model $g$ for this application:
\begin{enumerate}
    \item Sort the accepted students in non-decreasing order of 
    \[g(\mathbf{s}_i, \mathbf{g}_i, \$25,000|\theta) - g(\mathbf{s}_i, \mathbf{g}_i, \$0|\theta).\]
    \item Following the above order, allocate the maximum permissible aid (i.e., \$25,000) until reaching the maximum budget.
\end{enumerate}
This is a realistic heuristic because it greedily assigns scholarship to the students in the order of those that are most sensitive to scholarship.

Table~\ref{tab:heu} reports results from the experimental evaluation.  In particular, for logistic regression and neural network prediction models, we report for 500 and 1000 admitted students the expected number of enrolled students based on the allocation determined by $\Fworkname$ and the heuristic described above.  We also report for both models and for each $N$ the percent reduction in admitted student declination of admission.   The results indicate that simply by a more careful assignment of scholarship and making no other changes, $\Fworkname$ can provide a substantial improvement in expected matriculation rates.  This example exemplifies the improvement in decision-making capability that $\Fworkname$ can obtain.

\begin{table}[h!]
\footnotesize
\caption{Improvement by $\Fworkname$ over a basic heuristic.}
\label{tab:heu}
\begin{tabular}{c|c|c|c|c|c|c}
\centering
     & \multicolumn{3}{c|}{Logistic regression prediction} & \multicolumn{3}{c}{Neural network prediction}  \\ \hline
$N$  & $\Fworkname$    & Heuristic  & Exp. \% red. in declination  & $\Fworkname$    & Heuristic & Exp. \% red. in declination \\ \hline
500  & 445.02   & 437.59     & 11.91\%                    & 441.23 & 435.56    & 8.80\%                    \\
1000 & 887.02   & 872.26     & 11.55\%                    & 879.30 & 867.93    & 8.61\%                   
\end{tabular}
\end{table}
There are other ways that a practitioner who is well-versed in both optimization and predictive modeling might address a decision problem of this sort. For example, in this application, one could discretize the domain of the decision variables to take values $D = \set{0,1,\ldots,25,000}$ and then evaluate the predictive model $g(\cdot)$ at each value for each admitted student.  However, such a transformation results in a pseudo-polynomial size model and also admits only an approximation.  Note that if desired, one can model the problem in this way directly in $\Fworkname$ by declaring the decision variables as discrete and setting their domain to $D$.

\section{Accessing the Solver}
\label{sec:solveraccess}

$\Fworkname$ works with \texttt{Python3} and currently requires \href{https://www.gurobi.com}{\texttt{Gurobi}} for optimization and \href{https://scikit-learn.org/stable/}{\texttt{sklearn}} for predictive modeling.  You also must have \texttt{numpy} and \texttt{matplotlib} installed.
Please refer to $\Fworkname$'s website (\url{http://janos.opt-operations.com}) for more information, where a user manual, quick start guide, and examples are provided.

\section{Conclusions}
\label{sec:conclusion}

We propose a modeling framework $\Fworkname$ that integrates predictive modeling and prescriptive analytics. 
$\Fworkname$  is  a useful tool both for practitioners and researchers who are seeking to integrate machine learning models within a discrete optimization model.

%
%
%


\bibliographystyle{informs2014} 
\bibliography{solver} 


\end{document}